\documentclass{article}
\usepackage{ijcai23}
\usepackage{multirow}
\usepackage{amsfonts}
\usepackage{xcolor}
\usepackage{times}
\usepackage{soul}
\usepackage{url}
\usepackage[hidelinks]{hyperref}
\usepackage[utf8]{inputenc}
\usepackage[small]{caption}
\usepackage{graphicx}
\usepackage{amsmath}
\usepackage{amsthm}
\usepackage{booktabs}
\usepackage{algorithm}
\usepackage{algorithmic}
\usepackage[switch]{lineno}
\usepackage{mathtools}

\title{Sequence Learning Using Equilibrium Propagation}

\author{
Malyaban Bal
\and
Abhronil Sengupta
\affiliations
School of Electrical Engineering and Computer Science\\
The Pennsylvania State University\\
\emails
\{mjb7906, sengupta\}@psu.edu
}

\begin{document}

\maketitle

\begin{abstract}
    Equilibrium Propagation (EP) is a powerful and more bio-plausible alternative to conventional learning frameworks such as  backpropagation (BP). The effectiveness of EP stems from the fact that it relies only on local computations and requires solely one kind of computational unit during both of its training phases, thereby enabling greater applicability in domains such as bio-inspired neuromorphic computing. The dynamics of the model in EP is governed by an energy function and the internal states of the model consequently converge to a steady state following the state transition rules defined by the same. However, by definition, EP requires the input to the model (a convergent RNN) to be static in both the phases of training. Thus it is not possible to design a model for sequence classification using EP with an LSTM or GRU like architecture. In this paper, we leverage recent developments in modern hopfield networks to further understand energy based models and develop solutions for complex sequence classification tasks using EP while satisfying its convergence criteria and maintaining its theoretical similarities with recurrent BP. We explore the possibility of integrating modern hopfield networks as an attention mechanism with convergent RNN models used in EP, thereby extending its applicability for the first time on two different sequence classification tasks in natural language processing viz. sentiment analysis (IMDB dataset) and natural language inference (SNLI dataset). Our implementation source
code is available at \url{https://github.com/NeuroCompLab-psu/EqProp-SeqLearning}.
\end{abstract}

\section{Introduction}

Equilibrium Propagation (EP) \cite{scellier2017equilibrium} is a biologically plausible learning algorithm to train artificial neural networks. It requires only one computational circuit and single type of network unit during the two phases of training, whereas backpropagation (BP) requires a specialised type of computation during the backward phase to explicitly propagate errors which is different from the computational circuitry needed during the forward phase, thus it is essentially considered to be biologically implausible \cite{crick1989recent}. In EP, errors are propagated implicitly in the energy based model through local perturbations being generated at the output layer, unlike in BP. Moreover, strong theoretical connections \cite{scellier2019equivalence} between EP and recurrent BP \cite{almeida1990learning,pineda1987generalization} as well as the former's similarity regarding weight updates with spike-timing dependent plasticity (STDP) \cite{scellier2017equilibrium,bi1998synaptic} (a feasible model to understand synaptic plasticity change in neurons) makes EP a solid foundation to further understand biological learning \cite{lillicrap2020backpropagation}. Intrinsic properties of EP also provides the opportunity to design energy efficient implementations of the former on hardware unlike BP through time \cite{martin2021eqspike}. 

The idea that neurons collectively adjust themselves to configurations according to the sensory input being fed into a neural network system such that they can better predict the input data has been a popular hypothesis \cite{hinton2002training,berkes2011spontaneous}. The collective neuron states can be interpreted as explanations of the input data. EP also consists of this central idea where the network, which is essentially a dynamical system following certain dynamics, converges to lower energy states which better explains the static input data.

In this paper, we have primarily discussed EP in a discrete time setting and used the scalar primitive function $\phi$ \cite{ernoult2019updates,laborieux2021scaling} to derive the transition dynamics instead of an energy function as used in the initial works \cite{scellier2017equilibrium}. Primarily, algorithms using EP have been designed for convergent RNNs \cite{laborieux2021scaling} which are neural networks that take in a static input and through the recurrent dynamics (as governed by the transition function -- scalar primitive function of the system in our case) converges to a steady state which denotes the prediction of the network for that input. Training using EP primarily comprises of two distinct phases. During the first phase i.e. the ``free" phase, the network converges to a steady state following only the internal dynamics of the system. In contrast, in the second phase the output layer (which acts as the prediction after the ``free" phase has converged) is ``nudged" closer to the actual ground truth and the local perturbations resulting from that change propagates to the other layers of the convergent RNN, forming local error signals in time which matches the error propagation associated with BP through time \cite{ernoult2019updates}. The state updates and the consequent weight updates of the system can be done through local in space and time \cite{ernoult2019updates} computations, thus making learning algorithms using EP highly suitable for developing low-powered energy efficient neuromorphic implementations \cite{martin2021eqspike}. 

Hopfield networks were one of the earliest energy based models which were based on convergence to a steady state by minimizing an energy function which is an intrinsic attribute of the system. Hopfield networks were primarily used as associative memory, where we can store and retrieve patterns \cite{hopfield1982neural}. Classical hopfield networks were designed for storage and retrieval of binary patterns and the capacity of storage was limited since the energy function used was quadratic in nature. However, modern hopfield networks \cite{krotov2016dense,krotov2018dense,ramsauer2020hopfield} follow an exponential energy function for state transitions which allows storing exponential number of continuous patterns and allows for faster single step convergence \cite{demircigil2017model} during retrieval. The modern hopfield layer has been used previously for sequence-attention in conventional deep learning models with BP as the learning framework \cite{widrich2020modern}. In this paper, we integrate the capability of modern hopfield networks as a transformer-like \cite{vaswani2017attention} attention mechanism inside a convergent RNN and consequently train the resulting model using the learning rules defined by EP. 
\section{Motivation and Primary Contributions}
Until now, EP has been confined in the domain of image classification, the primary reason being the constraint associated with convergent RNNs which requires the input to be static. Developing an LSTM like network, which is fed with a time-varying sequence of data instead of static data, is not possible while maintaining the constraints of EP. Results obtained for tasks such as sentiment analysis or inference problems by feeding the entire input sequence as input also results in poor solutions since the dependencies between the different parts of the sequence cannot be captured in that technique. However, with recent developments in the domain of modern hopfield networks, we have the ability to extend the capability of algorithms using EP to perform efficient sequence classification. 

Modern hopfield networks fit perfectly with convergent RNNs since both of them are energy-based models which are governed by their respective state transition dynamics. In a discrete time setting - which is primarily discussed in this paper - both of them follow a certain state transition rule at every time step which is governed by their defining energy function or a scalar primitive function in case of a convergent RNN. As we will see in later sections, the one step convergence guarantee of modern hopfield networks allows us to seamlessly interface them with a convergent RNN allowing both the networks to settle to their respective equilibrium states as the system converges as a whole.

\noindent \textbf{Neuromorphic Motivation for EP:} The bio-plausible local learning framework that EP provides can be best utilised in a neuromorphic system. Spiking implementation of the proposed method following the works of \cite{martin2021eqspike} allows for low-powered solution for real-time scenarios such as intrusion detection, real-time natural language processing (NLP), etc. Implementing EP in a neuromorphic system allows for orders of magnitude more energy savings than in GPUs and is more efficient because of its single computational circuit and STDP like weight updates.

\noindent \textbf{Algorithmic and Theoretical contributions:} 
The primary contributions of our paper are as follows -
\begin{itemize}
\item We explore for the first time how hopfield networks can operate with a convergent RNN model and how we can leverage the attention mechanism of the former to solve complex sequence classification problems using EP.
  \item We illustrate mathematically and empirically how to combine the state transition dynamics of both the hopfield network and underlying convergent RNN to converge to steady states during both phases of EP thus maintaining the latter's theoretical equivalence with recurrent BP w.r.t. gradient estimation.
 
  \item We report for the first time the performance of EP as a learning framework on widely known NLP tasks such as sentiment analysis (IMDB dataset) and natural language inference (NLI) problems (SNLI dataset) and compare the results with state-of-the-art architectures for which neuromorphic implementations can be developed.
\end{itemize}
\section{Methods}
In this section, we will formulate the workings of EP and its theoretical connections with backpropagation through time (BPTT). We will then delve into the details of the modern hopfield layer and demonstrate its attention mechanism. We will elaborate on the scalar primitive function defined for state transition and further investigate the metastable states that arises in modern hopfield networks and will discuss how we can leverage them to form efficient encoding of the input sequences for our sequence classification problems.
\subsection{Equilibrium Propagation}
EP applies to convergent RNNs whose input at each time step is static. The state of the network eventually settles to a steady-state following a state update rule that is derived from the scalar primitive function $\phi$. Moreover, the weights defined between two layers are symmetric in nature i.e. if the weight between layers $s^i$ and $s^{i+1}$ is $w_i$, then the weight of the connection between $s^{i+1}$ and $s^i$ is $w_i^T$. The state transition is defined as,  
\begin{equation}
\label{eqn1}
s_{t+1} = \frac{\partial \phi}{\partial s}(x, s_t, \theta)
\end{equation}
where, $s_t = (s_t^1, s_t^2, \dots, s_t^n)$ is the collective state of the convergent RNN with $n$ layers at time $t$, $x$ is the input to the convergent RNN and $\theta$ represent the network parameters i.e. it comprises of the weights of each of the connections between layers. We do not consider any skip connection or self-connection in the convergent RNNs discussed but there has been some work done in that area \cite{gammell2021layer}. The state transitions result in a final convergence to a steady state $s_*$ after time $T$, such that $s_t = s_*$  $\forall t \geq T$ and it satisfies the following condition,
\begin{equation}
\label{eqn2}
s_{*} = \frac{\partial \phi}{\partial s}(x, s_*, \theta)
\end{equation}
Training of convergent RNNs using EP comprises mainly of two different phases. During the first phase or the ``free" phase, the RNN follows the transition function as shown in Eq. (\ref{eqn1}) and eventually reaches the steady state defined as the free fixed point $s_*$ after $T$ time steps. We use the output layer of the steady state $s_*$ i.e. $s^n_*$ to make the prediction for the current input $x$. In the second phase or the ``nudge" phase, an additional term $-\beta \frac{\partial L}{\partial s}$ is added to the state dynamics which immediately results in the state of the output layer being slightly nudged in the direction to minimize the loss function $L$ (as defined between the target $y$ and the output of the last layer of the convergent RNN). Though the internal states of the hidden layers are initially at the free fixed point state, they are eventually nudged to a different fixed point - weakly clamped fixed point $s_*^{\beta}$ - because of the perturbations initially originated at the output layer.  $\beta$ is a small scaling factor defined as the ``influence parameter" or ``clamping factor" i.e. it controls the influence of the loss on the actual primitive scalar function during the second phase.

Thus the initial state of the second phase is $s^\beta_0 = s_*$ and the transition function is defined as,
\begin{equation}
\label{eqn3}
s^{\beta}_{t+1} = \frac{\partial \phi}{\partial s}(x, s^{\beta}_t, \theta) -\beta \frac{\partial L}{\partial s}(s^{\beta}_t, y)
\end{equation}
Following Eq. (\ref{eqn3}), the convergent RNN settles to the steady state $s^{\beta}_*$ after $K$ timesteps. After the two phases are done, the learning rule to update the model parameters in order to minimize the loss $L^* = L(s_*, y)$ is defined as, $\Delta \theta = \eta \nabla^{EP}_\theta(\beta)$, \cite{scellier2017equilibrium} where $\eta$ is the learning rate and $\nabla^{EP}_\theta(\beta)$ can be defined as,
\begin{equation}
\label{eqn4}
\nabla^{EP}_\theta(\beta) = \frac{1}{\beta}(\frac{\partial \phi}{\partial \theta}(x, s^{\beta}_*, \theta) -\frac{\partial \phi}{\partial \theta}(x, s_*, \theta))
\end{equation}

The defined convergent RNN can also be trained by BPTT \cite{laborieux2021scaling}. According to the property of Gradient-Descent Updates (GDU) \cite{ernoult2019updates}, the gradient updates computed by the EP algorithm is approximately equal to the gradient computed by BPTT ($\nabla^{BPTT}(t)$), according to relative mean squared error metric (RMSE), provided the convergent RNN has reached its steady state in $T-K$ steps during the first phase and $\beta \rightarrow 0$. Thus for initial $K$ steps of the ``nudge" phase we can state ,
$\forall t = 1,2,...,K$
\begin{equation}
\label{eqn5}
\nabla^{EP}_\theta(t,\beta) = \frac{1}{\beta}(\frac{\partial \phi}{\partial \theta}(x, s^{\beta}_t, \theta) -\frac{\partial \phi}{\partial \theta}(x, s_*, \theta))
\end{equation}
\begin{equation}
\label{eqn6}
\nabla^{EP}_\theta(t,\beta) \xrightarrow[\beta \rightarrow 0]{} \nabla^{BPTT}(t)
\end{equation}
In the traditional implementations of EP, two phases are involved, one with $\beta = 0$ i.e. the ``free" phase and the other with $\beta > 0$ i.e. the ``nudge" phase. In order to circumnavigate the first order bias that is induced into the system by assuming $\beta > 0$, a new implementation of EP was proposed \cite{laborieux2021scaling} which comprises of a second ``nudge" phase with $-\beta$ as the influence factor. Thus the algorithm comprises of three phases. The symmetric EP gradient estimates are thus free from first order bias and are more close to  the values computed using BPTT and is defined as, 
\begin{equation}
\label{eqn7}
\nabla^{EPsym}_\theta(\beta) = \frac{1}{2\beta}(\frac{\partial \phi}{\partial \theta}(x, s^{\beta}_*, \theta) -\frac{\partial \phi}{\partial \theta}(x, s^{-\beta}_*, \theta))
\end{equation}
\subsection{Hopfield Network as Attention Mechanism}
The recent developments in modern hopfield networks  \cite{ramsauer2020hopfield} offers exponential storage capacity and one step retrieval of stored continuous patterns. The increased storage capacity and the attention-like state update rule of modern hopfield layers can be leveraged to retrieve complex representation of the stored patterns which consists of richer embedding similar to that of attention mechanism in transformers. In order to allow for continuous states, the energy function of the modern hopfield network is modified accordingly \cite{widrich2020modern}, and it can be represented as,
\begin{equation}
\label{eqn8}
E = -lse(\beta , X^T \xi) + \frac{1}{2} (\xi ^ T \xi) + \beta ^{-1} log N + \frac{1}{2} M^2
\end{equation}
\begin{displaymath}
lse(\beta , x) = \beta^{-1}log(\sum_{i=1}^Nexp(\beta x_i))
\end{displaymath}
where, $X = (x_1, x_2, ... x_N)$ are $N$ continuous stored patterns, $\xi$ is the state pattern, $M$ is the largest norm of all the stored patterns and $\beta > 0$. In general, the energy function of modern hopfield networks can be defined as, $E = -\sum_{i=1}^NF(x_i^T\xi)$ \cite{krotov2016dense}. For example, if we use the function $F(x) = x^2$, we describe the classical hopfield network which had limitations in storage capacity and also supported only binary patterns. However, with the introduction of an exponential interaction function like log-sum-exponential ($lse$) function, the storage capacity can be increased exponentially while enabling continuous patterns to be stored. Using Concave-Convex-Procedure (CCCP) \cite{yuille2001concave,yuille2003concave}, the state update rule for the modern hopfield network \cite{ramsauer2020hopfield} can be defined as,
\begin{equation}
\label{eqn9}
\xi^{new} = X\mathit{softmax}(\beta X^T \xi)
\end{equation}
where, $\xi^{new}$ is the retrieved pattern from the hopfield network. 

We can define Query ($Q = RW_Q$) as  $\xi ^ T$ and Key ($K = YW_K$) as $X^T$.  Thus the new form can be represented as, 
\begin{displaymath}
Q^{new} = \mathit{softmax}( \frac{1}{\sqrt{d_k}} RW_QW_K^TY^T)YW_K
\end{displaymath}

\begin{equation}
\label{eqn10}
Q^{new} = \mathit{softmax}( \frac{1}{\sqrt{d_k}} QK^T)K
\end{equation}
\begin{figure*}
  \centering
  \includegraphics[width=1.5\columnwidth]{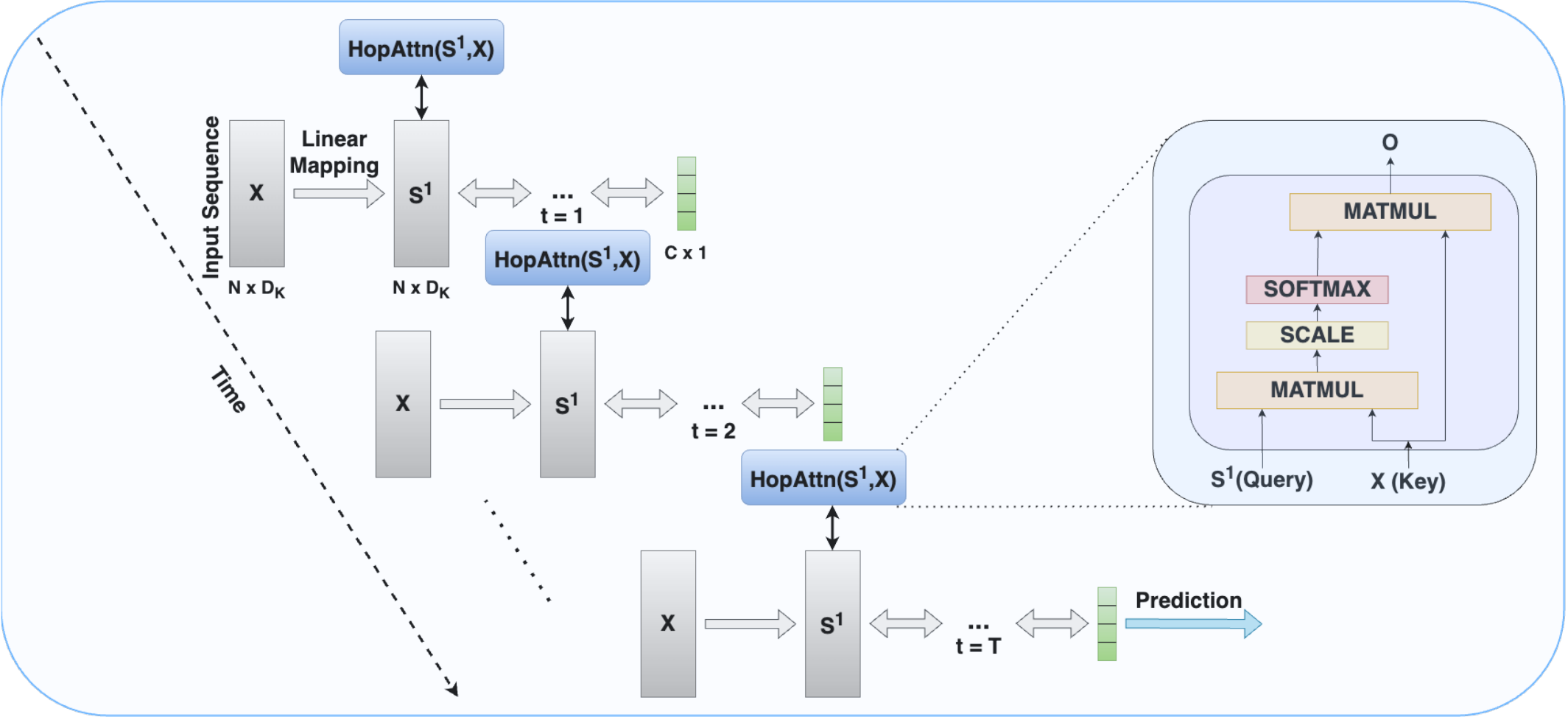}
  \caption{Simplistic scenario of application of $HopAttn$ modules in convergent RNN for a sequence classification problem with $c$ classes has been described. The input $X$ is directly fed as the Key $K$ to the $HopAttn$ module. During the ``free" phase (as described here), the model evolves through time following the defined state transition dynamics and converges to a steady state configuration after $T$ time steps to settle to the prediction. The operations inside the $HopAttn$ module are also illustrated in details.}
  \label{fig3}
\end{figure*}
where, $d_k$ is the encoding dimension of Key ($K$). The above representation \cite{ramsauer2020hopfield} is to show the similarity of the update rule of modern hopfield networks and the attention mechanism in transformers involving Query-Key pairs. $Q$ represents the state pattern whereas $Q^{new}$ represents the state pattern retrieved after the transition. The projection matrices are defined as, $W_Q \in \mathbb{R}^{d_r \times d_k}$ and $W_K \in \mathbb{R}^{d_y \times d_k}$.

Following the state transition as defined in Eq. (\ref{eqn10}), the Hopfield Attention module can be defined as $HopAttn(Q,K)$, which has two critical inputs, viz. the Query ($Q$) which can be interpreted as the state pattern and the Key ($K$) which can be considered as the stored pattern. The dimension of the hopfield space is $d_k$. The underlying operations are illustrated in details in Fig. \ref{fig3}. The output of the $HopAttn$ function is defined as, 
\begin{equation}
\label{eqn12}
\mathit{HopAttn}(Q,K) = softmax( \frac{1}{\sqrt{d_k}} QK^T)K
\end{equation}

$HopAttn$ module plays a critical role in our network by providing an attention based embedding of the input sequence resulting from their convergence to metastable states at every time step. In order to form the Query ($Q$) for the $HopAttn$ module, we define the layers connected to the input as projection layers (as demonstrated in Fig. \ref{fig3}) whose weight is analogous to $W_Q$. At every timestep $t$ during the state transitions in the convergent RNN, the recurrent dynamics of the convergent RNN computes the Query ($Q$) to be fed to the $HopAttn$ module as illustrated in Eq. (\ref{eqn14}). The stored sequence of patterns $K$ is fed directly into the $HopAttn$ module.

There is a theoretical guarantee that the modern hopfield network converges to a steady state (with exponentially small separation) after a single update step \cite{ramsauer2020hopfield} and therefore we successfully navigate to a fixed point at every time step in EP. We project the $HopAttn$ output through $W_V \in \mathbb{R}^{d_k \times d_v}$, analogous to the weight of the output connection,  to generate the final output. The $HopAttn$ module thus provides a rich representation of the stored patterns which helps in efficient sequence classification.

\subsection{Proposed Scalar Primitive function $\phi$}
The input to our model is the sequence of patterns, $x \in R^{N \times D_K}$, where $N$ is length of the sequence and $D_K$ is the encoding dimension of the pattern. The state of the system at time $t$ is $s_t = (s_t^0, s_t^1, s_t^2, \dots, s_t^n)$, where $n$ is the number of layers in the convergent RNN. $\theta$ comprises the list of parameters of the network. The scalar primitive function ($\phi$), which defines the state transition rules of the convergent RNN is,
\begin{equation}
\label{eqn11}
\phi(x, {s}, \theta) = {s^{1}} \bullet ( x \cdot w_{1}) + {s^{2}}^{T} \cdot w_{2} \cdot F(s^{1}) +  \sum_{i=2}^{n - 1} {s^{i+1}}^{T} \cdot w_{i+1} \cdot s^{i}
\end{equation}
where $\bullet$ represents Euclidean scalar product of two tensors with same dimensions, $( x \cdot w_{1} )$ represents the linear projection of $x$ through $w_1 \in R^{D_K \times D_K}$ and $w_{i+1}$ is the weight of the connection between $s^{i}$ and $s^{i+1}$. The flattened output of the first projection layer ($F(s^1)$, where $F$ is the flattening operation from $(A,B)$ to $(1,AB)$), is then fed into the next of the $n-1$ fully connected layers. In this section, we have primarily discussed $\phi$ for cases where we have only one input sequence like the sentiment analysis task. For multi-sequence tasks like NLI, we have described $\phi$ in the technical appendix.

\subsection{Transition Dynamics and Convergence}
Our methodology seamlessly interfaces modern hopfield networks with convergent RNNs, thus allowing sequence classification using EP. The state transition dynamics of the layers in the convergent RNN (except the last layer), where $HopAttn$ is not applied, is given by the following,
\begin{equation}
\label{eqn13}
s^{i}_{t+1} = \sigma (\frac{\partial \phi}{\partial s^{i}}(x, s_t, \theta))
\end{equation}
where, $\sigma$ is an activation function that restricts the range of the state variables to $[0,1]$.
However, for the layers where we apply the hopfield attention mechanism, the state transition function is updated as follows,
\begin{equation}
\label{eqn14}
s^{j}_{t+1} = \mathit{HopAttn}(\frac{\partial \phi}{\partial s^{j}}(x, s_t, \theta), K)
\end{equation}
where, $s^{j}_{t+1}$ is the final state of the $j^{th}$ layer (with $HopAttn$ applied) at time $t$+1 and $K$ is the sequence of stored patterns in the hopfield network which is usually the input $x$ to the network.

Thus for the layers where we apply attention, we first follow the transition function of the underlying convergent RNN as defined by $\frac{\partial \phi}{\partial s^{j}}(x, s_t, \theta)$ and then we follow the update rule as derived from the modern dense hopfield network. Since convergence to a steady state is guaranteed after a single update step, there is no additional overhead for the hopfield network to converge and thus its state transition rule is only applied once. 

\begin{figure}[h]
\centering
\includegraphics[width=\columnwidth]{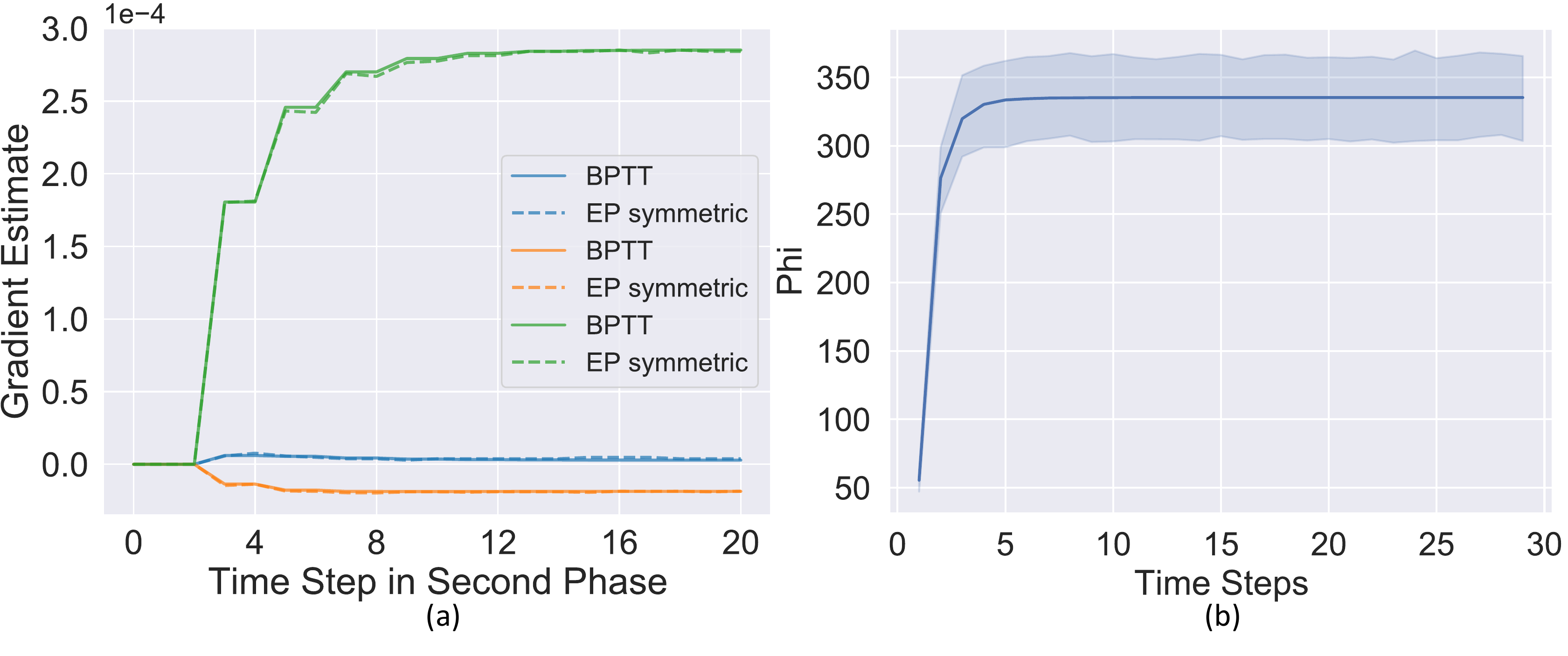}
\caption{Results taken from experiments run on IMDB dataset. (a) Symmetric EP gradient estimate as defined in Eq. (\ref{eqn7}) and gradient computed by BPTT, for three randomly chosen weights in the convergent RNN integrated with $HopAttn$ module. (b) Convergence of the scalar primitive function $\phi$ with time upon using $HopAttn$ in ``free" phase taken over 50 phase transitions.}
\label{fig2}
\end{figure}

The metastable states, that the $HopAttn$ module converges to after each timestep $t$, approach steady state w.r.t the convergent RNN during both the phases of training of EP. The claim for such concurrent convergence can be further substantiated empirically by analysing the convergence of the dynamics of the model, governed by the scalar primitive $\phi$ with time. It is evident from Fig. \ref{fig2}b that even after we apply the $HopAttn$ module (like in Fig. \ref{fig3}), the model still converges to a steady state eventually. Moreover, even after the integration of $HopAttn$ modules, the GDU property of the convergent RNN still holds, thus maintaining the equivalence w.r.t gradient estimation of EP and BPTT (Fig. \ref{fig2}a).

For the final layer, whose output is compared with the target label to calculate the loss function, the state update rule is defined as,
\begin{equation}
\label{eqn15}
s^{n}_{t+1} = \sigma (\frac{\partial \phi}{\partial s^n}(x, s_t, \theta)) + \beta (y - s^n_t)
\end{equation}
where, $y$ is the target label and $s^n_t$ is the output of the final layer at time $t$. $\beta = 0$ during the ``free phase" and the loss function $L$ used in this case is mean squared error.
\subsection{Metastable States as Attention Embeddings}
The factor $\beta$, as defined in Eq. (\ref{eqn9}), plays an important role in establishing the fixed points in the modern hopfield network \cite{ramsauer2020hopfield}. The retrieved state pattern (or $Q^{new}$ as defined above) settles to the defined fixed points following the state transition rule (Eq. (\ref{eqn10})). However, if $\beta$ is very high and/or the stored patterns are well separated, then we can easily retrieve the actual pattern stored in the hopfield networks. On the other hand, if $\beta$ is low and the stored patterns are not all well separated but form cluster like structures in the encoding dimension of the patterns, then the hopfield network generates metastable states. Thus, when we try to retrieve using a particular state pattern i.e a Query, we might converge to a metastable point comprising of a richer representation combining a number of patterns in that region. For incorporating such attention like behavior, we need to keep $\beta$ small and usually $\beta = 1/\sqrt{d_k}$, where $d_k$ is the encoding dimension of the stored patterns.
\subsection{Local State \& Parameter Updates}
The state update rule, as defined earlier, is essentially local in space since $\frac{\partial \phi}{\partial s^{i}}(x, s_t, \theta)$ $\forall i = 2,...,n-1$ can be written as,
\begin{equation}
\label{eqn16}
\frac{\partial \phi}{\partial s^i}(x, s_t, \theta) = w_i \cdot s_t^{i-1} + w_{i+1}^T \cdot s_t^{i+1}
\end{equation}
and for the final layer the same can be shown as,
\begin{equation}
\label{eqn17}
\frac{\partial \phi}{{\partial s^n}}(x, s_t, \theta) = w_n \cdot s_t^{n-1}
\end{equation}
where all the associated parameters for the computation are locally connected in space and time. The computation required w.r.t the $HopAttn$ module can also be done locally in space and time. For the first layer (projection layer), the local state and weight update property is still preserved: $\frac{\partial \phi}{\partial s^1}(x, s_t, \theta) = ( x \cdot w_{1} ) + F^{-1}(w_{2}^T \cdot s_t^{2}$).

The STDP like parameter update rule of the network, as stated in the earlier section, is also local in space, i.e. we can compute the updated weights directly using the state of the connecting layers. In this paper, we have used three phase EP, thus we derive the weight update rule following Eq. (\ref{eqn7}),
\begin{equation}
\label{eqn18}
\Delta w_{i+1} = \frac{1}{2\beta}(s^{i+1,\beta}_* \cdot s^{i,\beta ^T}_* - s^{i+1,-\beta}_* \cdot s^{i,-\beta ^T}_*)
\end{equation}
where, $s^{i,\beta}_*$ is the state of layer $i$ after the first ``nudge" phase with influence parameter $\beta$ and $s^{i,-\beta}_*$ is the state of layer $i$ after the second ``nudge" phase with influence parameter $-\beta$.


\begin{table*}
    \centering
    \begin{tabular}{| l | l | l | l | r |}
        \hline
           Task & Model & Method & Neuromorphic Implementation & Accuracy \\
         \hline
        \multirow{9}{4em}{Sentiment Analysis (IMDB)} & Simple Conv RNN & EP & Local opts.; No Non-diff. problem & $79.8 \pm 0.4$
        \\\cline{2-5}
        & ReLU GRU \cite{dey2017gate} &  \multirow{7}{1.5em}{BP}  &  \multirow{7}{14.5em}{Suffers gradient Non-diff. problem; Complex circuitry } & 84.8\\ 
         & GRU \cite{campos2017skip} & & & 86.5 \\ 
         & Skip GRU \cite{campos2017skip} & & & 86.6 \\
         & Skip LSTM \cite{campos2017skip} & & & 87.0 \\
         & LSTM \cite{campos2017skip} & & & 87.2 \\
         & CoRNN \cite{rusch2020coupled} & & & 87.4 \\
         & UniCORNN \cite{rusch2021unicornn} & & &88.4
        \\\cline{2-5}
         & \textbf{Our Model} & EP & Local opts.; No Non-diff problem. & \textbf{88.9} $\pm$ \textbf{0.3} \\
         \hline
         \multirow{7}{4em}{Natural Language Inference (SNLI)} &  100D LSTM encoders \cite{bowman2015large} & \multirow{6}{1.em}{BP} & \multirow{6}{14.5em}{Suffers gradient Non-diff. problem; Complex circuitry} & 77.6 \\ 
         & Lexicalized Classifier \cite{bowman2015large}  &  & & 78.2 \\
         & Parallel LMU \cite{chilkuri2021parallelizing} &  & & 78.8 \\ 
         & LSTM RNN encoders \cite{bowman2016fast} & & & 80.6 \\
         & DELTA - LSTM \cite{han2019delta} &  & & 80.7 \\
         & 300D SPINN-PI-NT \cite{bowman2016fast} &  & & 80.9
         \\\cline{2-5}
         & \textbf{Our Model} & EP & Local opts.; No Non-diff. problem & \textbf{81.4} $\pm$ \textbf{0.2} \\
         \hline
    \end{tabular}
    \caption{Comparing our models with other models trained using BP on the IMDB \& SNLI datasets.}
    \label{tab:booktabs}
\end{table*}

The detailed final algorithm and calculations are added in the technical appendix. It is further validated through the experiments reported in the next section that as we continue updating the states $s_t$ (following the update rules governed by the primitive function ($\phi$) and the energy function for the modern hopfield network), we reach steady states during both the ``free" and ``nudge" phases of EP. Thus, following the weight update procedure of EP, we can update the weights through local-in-space update rules. This enables us to develop state-of-the-art architectures for sentiment analysis and inference problems - potentially implementable in energy efficient neuromorphic systems.

\subsection{Neuromorphic Viewpoint}
Implementating EP in a neuromorphic setting reduces the energy consumption by two orders of magnitude during training and three orders during inference compared to GPUs \cite{martin2021eqspike}. Moreover, EP does not suffer from the non-differentiability of the spiking nonlinearity that BPTT algorithms encounter since we do not explicitly compute the error-gradient. In EP, because of the local operations and the error-propagation being spread across time, the memory overhead is also much lower compared to BPTT methods where we need to store the computational graph. 
\section{Experiments}
Since EP is still in a nascent stage, this is the first work to report the performance of convergent RNNs that are trained using EP on the specified datasets for the task of sequence classification. In the experiments reported in this section, we focus on benchmarking with models that are trained using BP such as LSTMs, GRUs, etc. that can be potentially implemented in a neuromorphic setting.  Unlike the used baselines, since existing attention models (transformers) trained using BP are not directly applicable in a neuromorphic setting; we have not compared our model with them. 
In the following sub-sections, we define specific architectures for different subtasks in NLP. For testing our proposed work on sentiment analysis problems, we chose the IMDB Dataset and for NLI problems, we chose the Stanford Natural Language Inference (SNLI) dataset. Though we have not tested our method on all the tasks in GLUE \cite{wang2018glue}, the proposed method can perform classification after applying simple task-specific adjustments to the model. Details regarding coding platform and hardware used for training have been added to the technical appendix.

\subsection{Sentiment Analysis}
We have used IMDB dataset \cite{maas2011learning} to demonstrate the application of our model on sentiment analysis tasks. IMDB dataset comprises of 50K reviews, 25K for training and 25K for testing. Each of the reviews are either classified as positive or negative. 
\subsubsection{Architecture} 300D word2vec embeddings \cite{mikolov2013efficient} are used for generating the word embeddings that are then fed into the convergent RNN and the maximum sequence length is restricted to 600. The convergent RNN used for this experiment comprises of two fully connected hidden layers on top of the first projection layer. We have a single attention module applied to the first layer similar to Fig. \ref{fig3}.
\subsubsection{Results} The results (Table \ref{tab:booktabs}) from our model are the first to report performance on any sentiment analysis task using EP. The experimental details are shown in Table \ref{table_imdb}.

\begin{table}[h!]
    \centering
\begin{tabular}{| l | l | r | r |} 
 \hline  
 Hyper-params \& Perf. &  Range & Optimal \\ 
 \hline
  Influence Factor ($\beta$) & (0.01-0.99) & 0.1 \\
  $T$ (``Free Phase") & (40-200) & 50\\ 
  $K$ (``Nudge Phase") &  (15-80) & 25 \\
  Epochs & (10-100) & 40 \\
  Layers (Linear \& FC) & - & (1 \& 3) \\
  Layer-wise lr & - &1e-4,5e-5,5e-5,5e-5 \\
  Batch Size & (8-512) &  128 \\
  \hline
  \hline
  Memory (GB) & - &  7 (128 batch size) \\
  Inference Time (sec) & - &  120 (Test Set) \\
 \hline
\end{tabular}
\caption{Hyper-parameters \& Perf. Metrics for IMDB dataset.}
\label{table_imdb}
\end{table}

\begin{figure}[h]
\centering
\includegraphics[width=\columnwidth]{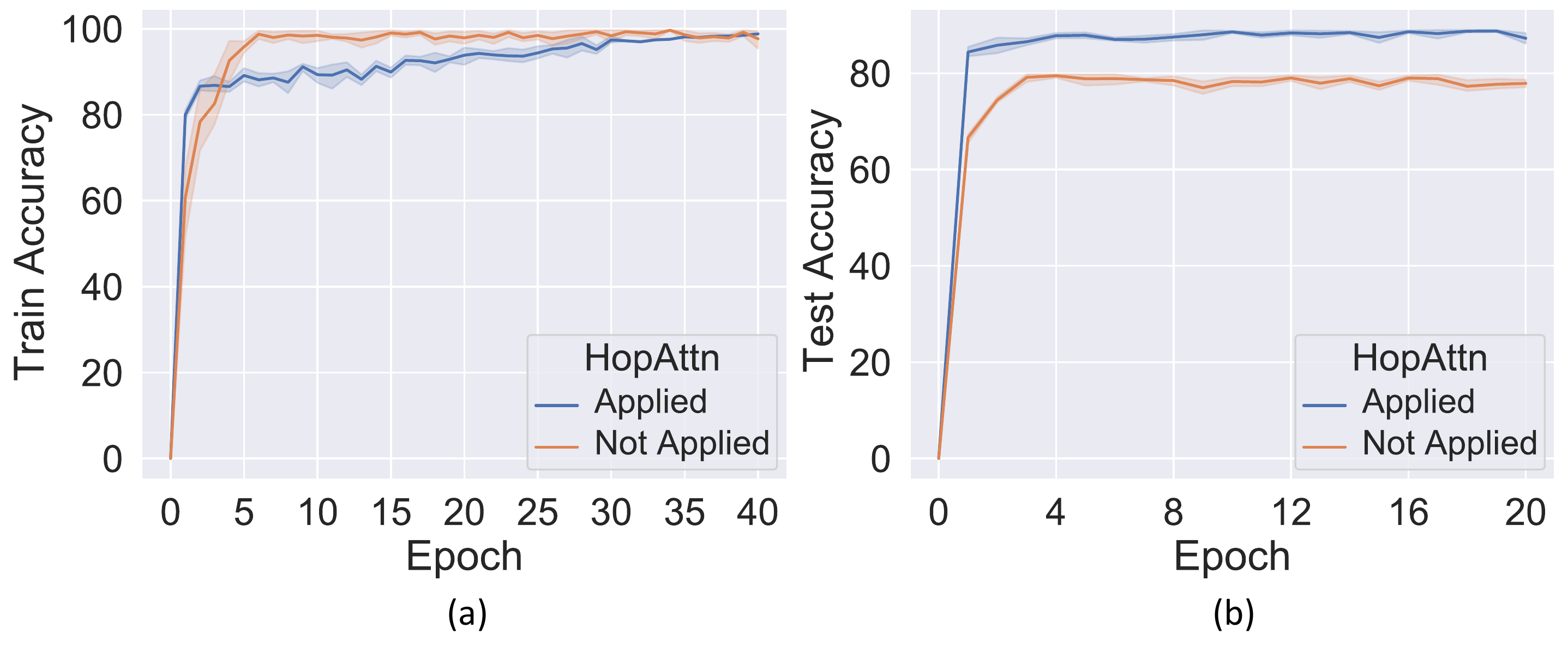}
\caption{(a) Train Accuracy and (b) Test Accuracy comparison of two different convergent RNN models trained with EP - one with a $HopAttn$ module applied (as shown in Fig. \ref{fig3})  and one without it. Results are reported on IMDB dataset over 5 different runs.}
\label{fig4}
\end{figure}

In order to study the advantage of using hopfield networks as an attention mechanism, we compare the accuracy achieved using our model against a vanilla implementation of EP on a convergent RNN with same depth but without any $HopAttn$ modules and we see that our proposed model outperforms with a big margin ( $ >9\%$).  The computational cost increases by only ~17\% when we use the $HopAttn$ module. The case where hopfield attention modules were not used seems to converge early during training, thereby resulting in over-fitting. However, as is evident from Fig. \ref{fig4}, usage of modern hopfield networks results in better generalization. 
\subsection{Natural Language Inference (NLI) Problems}
We have used SNLI \cite{bowman2015large} dataset to evaluate our model on NLI tasks. NLI generally deals with the classification of a pair of sentences namely, a premise and a hypothesis. A model, given both the sentences, is able to predict whether the relationship between the two sentences signify an entailment, contradiction or if they are neutral. SNLI dataset comprises of 570K pairs of premises and hypotheses.

\begin{table}[h!]
    \centering
\begin{tabular}{| l | l | r | r |} 
 \hline  
 Hyper-params \& Perf. &  Range & Optimal \\ 
 \hline
  Influence Factor ($\beta$) & (0.01-0.99) & 0.5 \\
  $T$ (``Free Phase") & (40-200) & 60\\ 
  $K$ (``Nudge Phase") &  (15-100) & 30 \\
  Epochs & (10-100) & 50 \\
  Layers (Linear \& FC) & - & (4 \& 2) \\
  Layer-wise lr & - & 4 * 5e-4,2e-4,2e-4 \\
  Batch Size & (8-1024) &  256 \\
  \hline
  \hline
  Memory (GB) & - & 1.9 (256 batch size) \\
  Inference Time (sec) & - &  ~40 (Test Set) \\
 \hline
\end{tabular}
\caption{Hyper-parameters \& Perf. Metrics for SNLI dataset.}
\label{table_snli}
\end{table}
\subsubsection{Architecture} 
\begin{figure}[h]
\label{snlifig}
\includegraphics[width=\columnwidth]{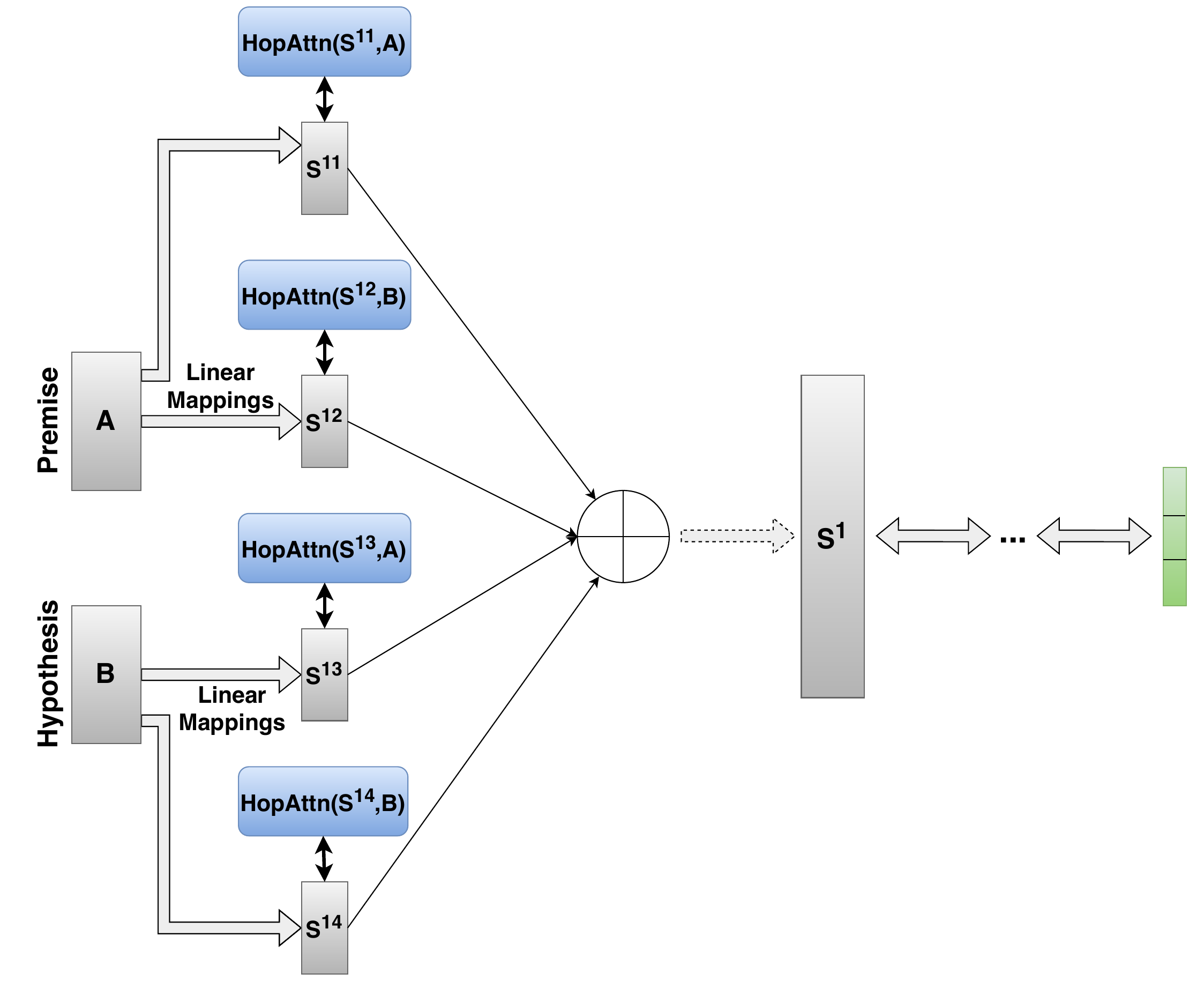}
\caption{High-level overview of the architecture used in case of SNLI dataset. $HopAttn$ modules are used to capture dependencies within different parts of a text as well as cross-attention between two separate texts. The output of each of the layers are concatenated. The network converges over time during both phases of EP.}
\label{fig6}
\end{figure}

The premise and the hypothesis are encoded as a sequence of 300D word2vec word embeddings \cite{mikolov2013efficient} with max sequence length of 25. The hopfield attention modules are used as specified in Fig. \ref{fig6}. The architecture defined for this problem is inspired from the decomposable  attention model \cite{parikh2016decomposable}. We denote the premise as $A = (a_1, a_2,...,a_m)$ and hypothesis as $B = (b_1, b_2,...,b_n)$, where $a_i,b_j \in R^d$, $d$ is the dimensionality of the word vector and $m$ and $n$ are the length of the sequences of the premise and hypothesis. We compute the new vectors $A^{'} = HopAttn(s^{11}, A)$ and  $B^{'} = HopAttn(s^{14}, B)$, where $s^{11}$ and $s^{14}$ are values of the layers as shown in Fig. \ref{fig6}. These two vectors represent self-attention within $A$ and $B$ respectively. We compute two other vectors, $\alpha = HopAttn(s^{13}, A)$ which encode the soft alignment of premise with the hypothesis $B$ and $\beta = HopAttn(s^{12}, B)$ which represents the soft alignment of hypothesis with the premise $A$, where $s^{13}$ and $s^{12}$ are values of the layers (see Fig. \ref{fig6}). We then concatenate all the sequences and feed them to the next layer (see Fig. \ref{fig6}).

\subsubsection{Results} The results obtained by our model (Table \ref{tab:booktabs})  are compared with other models trained using BP reported in literature that can be potentially implemented in a neuromorphic setting. 
The results from our model are the first to report performance on any NLI task using EP as a learning framework. The experimental details are shown in Table \ref{table_snli}.
\section{Conclusion}
The ability to think of neural networks as a dynamical system helps to further deepen our understanding regarding
learning frameworks and intrigues us to delve deeper into
the learning processes inside the brain. In this paper, we explore the application of EP as a learning framework in convergent RNNs integrated with modern hopfield networks to solve sequence classification problems. We report for the first time the performance of EP on datasets such as IMDB and SNLI. The constraint of EP requiring static input to the convergent RNNs makes it really difficult to train on datasets with sequence of data, thus an attention-like-mechanism provided by modern hopfield networks is ideal to encode the long-term dependencies in the sequence. The spatially (and potentially temporal) local weight update feature of EP still holds even after introducing the modern hopfield networks and therefore it can be easily converted into a neuromorphic implementation following the works of \cite{martin2021eqspike}.

Certain unexplored areas in the paper can be investigated in the future. Firstly, due to limitations of EP, we employed fixed word-embeddings instead of learnable-embeddings used in sophisticated language models. Thus circumnavigating that challenge to achieve even better accuracy can be an interesting problem. Secondly, another intrinsic restriction of the convergent RNN model used in the experiments is that the weights of the connections needs to be symmetric. In order to make the model more bio-plausible with asymmetric connections, the Vector Field \cite{scellier2018generalization} can be further explored. Finally, although storing small sequence lengths like that of SNLI is not a big concern, the memory overhead increases with increased sequence sizes like that of IMDB dataset. Thus a future endeavor can be made to modify EP such that it supports time-varying inputs.

\section*{Acknowledgments}
This material is based upon work supported in part by the U.S. Department of Energy, Office of Science, Office of Advanced Scientific Computing Research, under Award Number \#DE-SC0021562 and the National Science Foundation grant CCF \#1955815.

\clearpage
\section{Appendix: Algorithm}
The pseudocode for the algorithm designed to integrate modern hopfield networks into convergent RNNs and subsequently train the latter using the three phase (one free phase and two nudge phases) equilibrium propagation is described in this section. We have also elaborated on the scalar primitive function used for natural language inference task.

\subsection{Pseudocode}
\begin{algorithm}
\caption{($stateUpdate$) The state update dynamics of the convergent RNN integrated with $HopAttn$ module. Activation function $\sigma$ removed for simplicity and we consider the input $x$ as the stored patterns in the hopfield network.}
\label{alg:algorithm1}
\textbf{Input}: $x, y, s,$ $\theta$, $\beta$, $T_{\beta}$\\
\textbf{Parameter}: Optional list of parameters\\
\textbf{Output}: $s$
\begin{algorithmic}[1] 
\STATE $s_0 \gets s$
\FOR{$t = 0$ to $T_{\beta}$}
\FOR{$i = 1$ to $n-1$}
\STATE $s^i_{t+1} \gets \frac{\partial \phi}{\partial s^i}(x, s_t, \theta)$
\IF {$isHopAttnApplied(s^i_{t+1})$ is True}
\STATE $s^i_{t+1} \gets HopAttn(s^i_{t+1}, x)$
\ENDIF
\ENDFOR
\STATE $s^{n}_{t+1} \gets \frac{\partial \phi}{\partial s^n}(x, s_t, \theta) + \beta (y - s^n_{t})$
\ENDFOR
\STATE \textbf{return} $s_{T_\beta}$
\end{algorithmic}
\end{algorithm}

\begin{algorithm}
\caption{Three-phase EP applied on the proposed convergent RNN model integrated with $HopAttn$ layer.}
\label{alg:algorithm2}
\textbf{Input}: $x, y,$ $\theta$, $\eta$\\
\textbf{Output}: $\theta$
\begin{algorithmic}[1] 
\STATE $s_0 \gets 0$
\STATE $s_T \gets stateUpdate(x,y,s_0, \theta, 0, T)$ 
\STATE $s_* \gets s_T$
\STATE $s^{\beta}_* \gets stateUpdate(x,y,s_*, \theta, \beta, K)$
\STATE $s^{-\beta}_* \gets stateUpdate(x,y,s_*, \theta, -\beta, K)$
\STATE $\nabla^{EPsym}_\theta \gets \frac{1}{2\beta}(\frac{\partial \phi}{\partial \theta}(x, s^{\beta}_*, \theta) -\frac{\partial \phi}{\partial \theta}(x, s^{-\beta}_*, \theta))$
\STATE $\theta \gets \theta + \eta \nabla^{EP}_\theta$
\STATE \textbf{return} $\theta$
\end{algorithmic}
\end{algorithm}
\subsection{Scalar Primitive Function $\phi$ - SNLI}
The scalar primitive function $\phi$ is used to derive the state transition dynamics of the convergent RNN. In the paper, a high level overview of the function $\phi$ is described where the task described has only one sequence as an input. However in order to understand the sequence classification problems with multiple inputs $x_1, x_2 \in R^{N \times D_K}$, we need to further elaborate few underlying operations. If we consider the natural language inference problem (SNLI dataset) with the projection layers as defined in Fig. \ref{fig6}, we can define the $\phi$ as,
\begin{multline}
\label{eqn19}
\phi(x, {s}, \theta) = \sum_{j=1}^{2} {s^{1j}}  \bullet ( x_1 \cdot w_{1j}) + \sum_{j=3}^{4} {s^{1j}}  \bullet ( x_2 \cdot w_{1j}) \\ +{s^{2}}^{T} \cdot w_{2} \cdot F(s^{1}) + \sum_{i=2}^{n - 1} {s^{i+1}}^{T} \cdot w_{i+1} \cdot s^{i}
\end{multline}
where $\bullet$ represents euclidean scalar product of two tensors with same dimensions, $( x_i \cdot w_{1j} )$ represents the linear projection of $x_i$ through $w_{1j} \in R^{D_K \times D_K}$, $\forall j = 1,2,...,4$, and $w_{i+1}$ is the weight of the connection between $s^{i}$ and $s^{i+1}$, $\forall i = 1,...,n-1$. $s^1$ is the concatenation of all four projection layers $s^{1j}$, $\forall j = 1,2,...,4$. The flattened output of the concatenation of all four projection layers ($F(s^1)$ where $F$ is the flattening operation from $(A,B)$ to $(1,AB)$) is then fed into rest of the $n-1$ fully connected layers.
\section{Appendix: Experimental Details}
We implemented modified EP, as given by Algorithm \ref{alg:algorithm2}, using PyTorch 1.7.0. The experiments were run on Nvidia GeForce RTX 2080 Ti GPUs (8). The pre-processed  IMDB dataset available in Keras is used in our experiments.

\subsection{Sentiment Analysis (IMDB dataset)}

The pre-processed  IMDB dataset, available in Keras, is used in our experiments. In the sentiment analysis problems, we have 300D word2vec embeddings with maximum sequence length of 600 as the input to the convergent RNN. The first layer is a linear projection layer which maintains the same encoding dimension of 300. The hopfield attention module is integrated with the first layer. The final output of that layer is flattened and fed into a hidden layer of size 1000, followed by another layer of size 40 and then followed by the output layer of size 2. Adam optimizer is used as the optimizer to update the weights and the activation function is a modified form of hard sigmoid function. The loss function, as discussed in the paper, is mean squared error. GPU memory usage while training is limited to around 7GB with batch size of 128 and running time for one experiment is around 5 hours on the described hardware. 

\subsection{Natural Language Inference (SNLI dataset)}

We have 300D word2vec embeddings with maximum sequence length of 25 for both premise and hypothesis given as the input. The scalar primitive function can be considered to be of a similar form as given by Eqn. \ref{eqn19}. However, we consider four linear projection layers connected to inputs $A$ and $B$, as depicted in the paper. Then the output of all the projection layers are concatenated and flattened to be fed into the remaining two fully connected layers. The architecture is kept simple to show the effectiveness of using the modern hopfield network. Activation function and optimizer used is the same as the previous experiment.  GPU memory usage during training is limited to around 1.9 GB with batch size of 256. Running time for one experiment is around 6 hours.

\subsection{Implementation Details}
In order to make the code simpler and intuitive, the automatic differentiation framework of PyTorch has been used in our code to calculate the state and weight update rules. The scalar primitive function $\phi$ has been defined for each of our test datasets according to the underlying architecture and then the state update dynamics and weight updates have been computed by taking the appropriate derivatives of $\phi$. Such an implementation also makes the code robust to further changes in architecture. 

\end{document}